%% file: main.tex
\documentclass[10pt,twocolumn,letterpaper]{article}

\usepackage{cvpr}              
\usepackage[utf8]{inputenc}
\usepackage[T1]{fontenc}

\usepackage{marvosym}
\usepackage{booktabs} 
\usepackage{amsmath,amsfonts}
\usepackage{algorithmic}
\usepackage{algorithm}
\usepackage{array}
\usepackage{textcomp}
\usepackage{overpic}
\usepackage{stfloats}
\usepackage{url}
\usepackage{xspace}
\usepackage{verbatim}
\usepackage{graphicx}
\usepackage{subcaption}
\usepackage{listings}
\usepackage{tabularx}
\usepackage{xcolor}  
\usepackage{colortbl}
\usepackage{pifont}
\usepackage{amssymb}
\definecolor{cell}{gray}{0.90} 

\usepackage{multirow}

\usepackage{soul}
\definecolor{frenchblue}{rgb}{0.0, 0.45, 0.73}
\input{preamble}
\definecolor{cvprblue}{rgb}{0.21,0.49,0.74}
\usepackage[pagebackref,breaklinks,colorlinks,allcolors=cvprblue]{hyperref}

\def\paperID{773} 
\def\confName{CVPR}
\def\confYear{2026}

\title{Unlocking 3D Affordance Segmentation with 2D Semantic Knowledge}

\author{
Yu Huang$^{1}$, Zelin Peng$^{1,\dag}$, Changsong Wen$^{1}$, Xiaokang Yang$^{1}$, and Wei Shen$^{1{(\textrm{\Letter})}}$\\
$^1$MoE Key Lab of Artificial Intelligence, School of Computer Science, Shanghai Jiaotong University\\
{\tt\small \{yellowfish, zelin.peng, changsong, xkyang, wei.shen\}@sjtu.edu.cn}\\
}
\begin{document}
\twocolumn[{%
\renewcommand\twocolumn[1][]{#1}%
\maketitle
\vspace{-2.2em}

\begin{center}
    \begin{minipage}{1.0\textwidth}
        \centering
        \includegraphics[width=\linewidth,keepaspectratio]{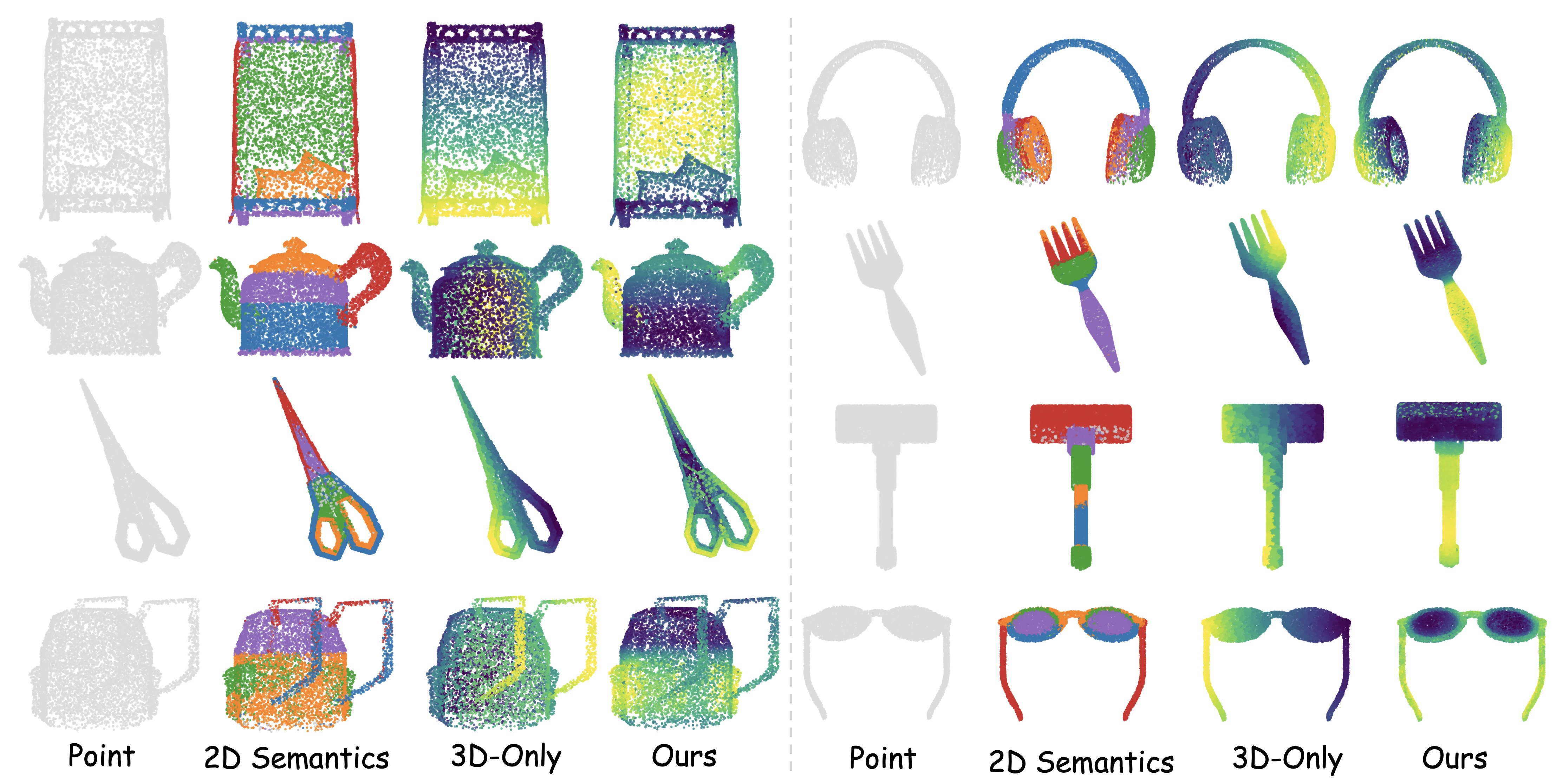}
        \vspace{-7mm}
        \captionof{figure}{\textbf{Qualitative comparison of 3D feature representations.} We visualize learned features across different objects to examine how semantic structure emerges in 3D space. In the \textbf{2D Semantics} column, lifted features from multi-view renderings encoded by a 2D vision foundation model (e.g., DINOv3~\cite{dinov3}) reveal clear functional clusters such as handles and seats. The \textbf{3D-Only} column, derived from a pure 3D encoder (e.g., PointNet++~\cite{qi2017pointnetplusplus}), shows less organized patterns, where boundaries between object parts remain fuzzy and inconsistent. In contrast, the \textbf{Ours} column shows features that inherit 2D semantic organization and express it coherently in 3D. Functional parts become more distinctly separated, and similar regions stay consistent across different object categories.
        }
        \label{figure1}
        \vspace{-2.5mm}
    \end{minipage}
\end{center}
\vspace{1em}
}]

\newcommand\blfootnote[1]{%
\begingroup 
\renewcommand\thefootnote{}\footnote{#1}%
\addtocounter{footnote}{-1}%
\endgroup 
}
\blfootnote{$^{\textrm{\Letter}}$ Corresponding Author: \texttt{wei.shen@sjtu.edu.cn}}
\blfootnote{$^\dag$ Project Leader.}

\input{sec/0_abstract}    
\input{sec/1_intro}

\input{sec/2_related}
\input{sec/3_method}
\input{sec/4_experiment}
\input{sec/5_conclusion}

\clearpage
{
    \small
    \bibliographystyle{ieeenat_fullname}
    \bibliography{main}
}


\end{document}

%% file: sec/0_abstract.tex

\begin{abstract}
Affordance segmentation aims to decompose 3D objects into parts that serve distinct functional roles, enabling models to reason about object interactions rather than mere recognition. Existing methods, mostly following the paradigm of 3D semantic segmentation or prompt-based frameworks, struggle when geometric cues are weak or ambiguous, as sparse point clouds provide limited functional information. To overcome this limitation, we leverage the rich semantic knowledge embedded in large-scale 2D Vision Foundation Models (VFMs) to guide 3D representation learning through a cross-modal alignment mechanism. Specifically, we propose Cross-Modal Affinity Transfer (CMAT), a pretraining strategy that compels the 3D encoder to align with the semantic structures induced by lifted 2D features. CMAT is driven by a core affinity alignment objective, supported by two auxiliary losses, geometric reconstruction and feature diversity, which together encourage structured and discriminative feature learning. Built upon the CMAT-pretrained backbone, we employ a lightweight affordance segmentor that injects text or visual prompts into the learned 3D space through an efficient cross-attention interface, enabling dense and prompt-aware affordance prediction while preserving the semantic organization established during pretraining. Extensive experiments demonstrate consistent improvements over previous state-of-the-art methods in both accuracy and efficiency.
\end{abstract}

%% file: sec/1_intro.tex
\section{Introduction} \label{sec:intro}

Affordance segmentation focuses on dividing a 3D object into parts that carry distinct functional roles. For example, a chair can be separated into a seat, a backrest, and legs. By identifying such functional components, intelligent systems can move beyond passive object recognition and start learning about how to interact with the object in purposeful ways. Early approaches predominantly followed the pipeline of 3D semantic segmentation~\cite{graham20183d,tchapmi2017segcloud,rozenberszki2022language}, where point cloud encoders predict part-level labels based solely on geometric information. This paradigm assumes that functional distinctions can be inferred directly from shape. However, many affordances are not uniquely determined by local geometry: the graspable handle of a mug can be geometrically similar to its rim, and surfaces that afford support or contact often exhibit smooth or symmetric forms. When geometric cues are weak or ambiguous, especially under sparse scanning, occlusion, or noisy reconstruction, these models tend to produce unstable or coarse functional boundaries. These observations indicate that geometric structure alone is insufficient to capture part-level semantics.

To compensate for missing functional semantics, recent methods introduce prompt-based affordance segmentation, where visual demonstrations or textual instructions guide the prediction process~\cite{great,LASO,geal,zhu2025grounding3dobjectaffordance}. For example, a textual query such as “Where should this mug be grasped?” or a visual prompt showing a hand-holding posture can highlight the relevant functional regions. Building on this idea, some systems further incorporate multimodal large language models (MLLMs)~\cite{wu2024visionllmv2,internvl,llava,cot} to make prompt interpretation more flexible and expressive. However, even with improved prompt processing, these approaches often yield constrained improvements relative to their added complexity. We believe this may point to a deeper issue: their performance remains suboptimal partly because they still rely on a 3D encoder trained primarily as a geometric feature extractor. This suggests that the bottleneck lies not in the prompts themselves, but in the representational capacity of the encoder. Sparse point clouds inherently contain limited functional cues, and without a feature space that encodes semantic-aware structure, prompts cannot reliably impose such semantics. Thus, affordance segmentation requires rethinking how 3D features are learned, rather than simply enriching the prompting modalities.

Since the core bottleneck lies in the 3D encoder’s semantic capacity, a natural question arises: how can we inject stronger semantic structure into 3D features? One promising path is to draw on large-scale 2D Vision Foundation Models (VFMs)~\cite{clip,dinov1,dinov2,dinov3}. These models, which are trained on extensive image corpora without supervision, naturally learn feature spaces that inherently capture clear and structured semantic organization. As shown in Fig.~\ref{figure1} (2D Semantics), features derived from models such as DINOv3~\cite{dinov3} already form clusters that correspond closely to functionally coherent regions of objects. This indicates that 2D visual representations naturally encode semantic-aware structure more readily than purely geometric 3D embeddings. Motivated by this, recent works~\cite{uad,d3field,locate3d,conceptfusion} seek to transfer such semantic knowledge into the 3D domain by “lifting” multi-view 2D features onto point clouds. The lifted features can serve as dense semantic supervision signals, guiding the 3D encoder to develop a well-structured feature space. 

Building on this insight, we propose a novel learning paradigm for affordance segmentation. Our approach begins by distilling semantic knowledge from a VFM into the 3D domain through multi-view feature lifting, thereby generating dense, per-point semantic-aware guidance. To ensure that our 3D encoder effectively internalizes this knowledge, we introduce Cross-Modal Affinity Transfer (CMAT), a novel pretraining strategy that compels the encoder to align with the semantic structures induced by the lifted 2D features. By optimizing our core affinity alignment objective, which is supported by two auxiliary losses, namely geometric reconstruction and feature diversity, CMAT builds a 3D backbone capable of producing highly structured and discriminative features.

We then deploy our CMAT-pretrained backbone within a lightweight affordance segmentor (LAS), which serves as an efficient architecture for the task while simultaneously validating our backbone's effectiveness. Unlike previous prompt-driven pipelines that depend on large multimodal language models to interpret interaction cues, our segmentor employs a lightweight cross-attention interface that injects text or visual prompts directly into the CMAT-pretrained 3D feature space. This simple-but-effective design avoids redundant reasoning modules and preserves the clean, structured semantics learned during pretraining. Empirically, this compact architecture supports dense, prompt-aware affordance segmentation with less computational overhead, and consistently achieves a clear performance margin over state-of-the-art methods.


%% file: sec/2_related.tex
\section{Related Work}
\noindent\textbf{Affordance Segmentation in 3D.}
Modern approaches to 3D affordance segmentation~\cite{Luo_2023_CVPR, Oneluo, zhai2023background, luo2021one,luotnnls,li:ooal:2024,partafford} primarily rely on deep neural networks trained on fully annotated datasets. These methods have become adept at learning the correspondence between local geometric patterns in point clouds and their associated functional labels. However, their performance is fundamentally tied to the quality and scale of 3D supervision, often struggling to generalize to unseen object categories. A core limitation is their reliance on geometric cues alone, which can be ambiguous; for instance, a flat surface could be a “sittable” seat or a “supportable” tabletop, a distinction that requires contextual semantic reasoning beyond local shape.

\noindent\textbf{Multi-modal Guidance for 3D Affordance Segmentation.}
To enhance semantic reasoning, a dominant trend~\cite{Yang_2023_ICCV,LASO,Nguyen2023open,van2023open, yang2024egochoir,zhaipami,guo2026} is the use of multi-modal guidance, particularly from Vision-Language Models (VLMs) ~\cite{wu2024visionllmv2,internvl,llava,cot,star,peng2025parameter,article} like CLIP~\cite{clip}. Beyond CLIP-based alignment, OpenAD~\cite{openad} and OpenKD~\cite{openkd} further explore text–point correlation and synonym substitution for open-vocabulary affordance grounding \cite{openad,openkd}. These prompt-driven architectures enable remarkable zero-shot and open-vocabulary segmentation by aligning point cloud features with textual or visual prompts. While these models demonstrate impressive flexibility, their success still depends heavily on the representational quality of the underlying 3D encoder, which may lack fine-grained discriminability to separate functionally distinct parts.

\noindent\textbf{Knowledge Transfer from 2D VFMs to 3D.}
A promising direction for improving 3D representations is transferring semantic knowledge from large-scale 2D Vision Foundation Models (VFMs). A prevalent technique~\cite{uad} is to lift features from multi-view images extracted by models such as DINO and project them into 3D space, enriching point cloud representations with semantic cues that raw geometry cannot provide. Prior studies~\cite{d3field,locate3d,conceptfusion} show that these lifted features inject strong semantic signals, helping to impose organization on otherwise ambiguous point clouds and leading to more discriminative feature spaces. However, most existing work focuses on aligning individual features or ensuring broad consistency between 2D and 3D semantics, without explicitly modeling the relational structure among parts, which can leave representations fragmented or inconsistent. Motivated by these limitations, our work explores a semantic-grounded approach that more explicitly structures 3D representations, aiming to achieve finer part-level discrimination required for affordance segmentation.

%% file: sec/3_method.tex
\begin{figure*}[ht]
    \centering
    \includegraphics[width=1.0\linewidth]{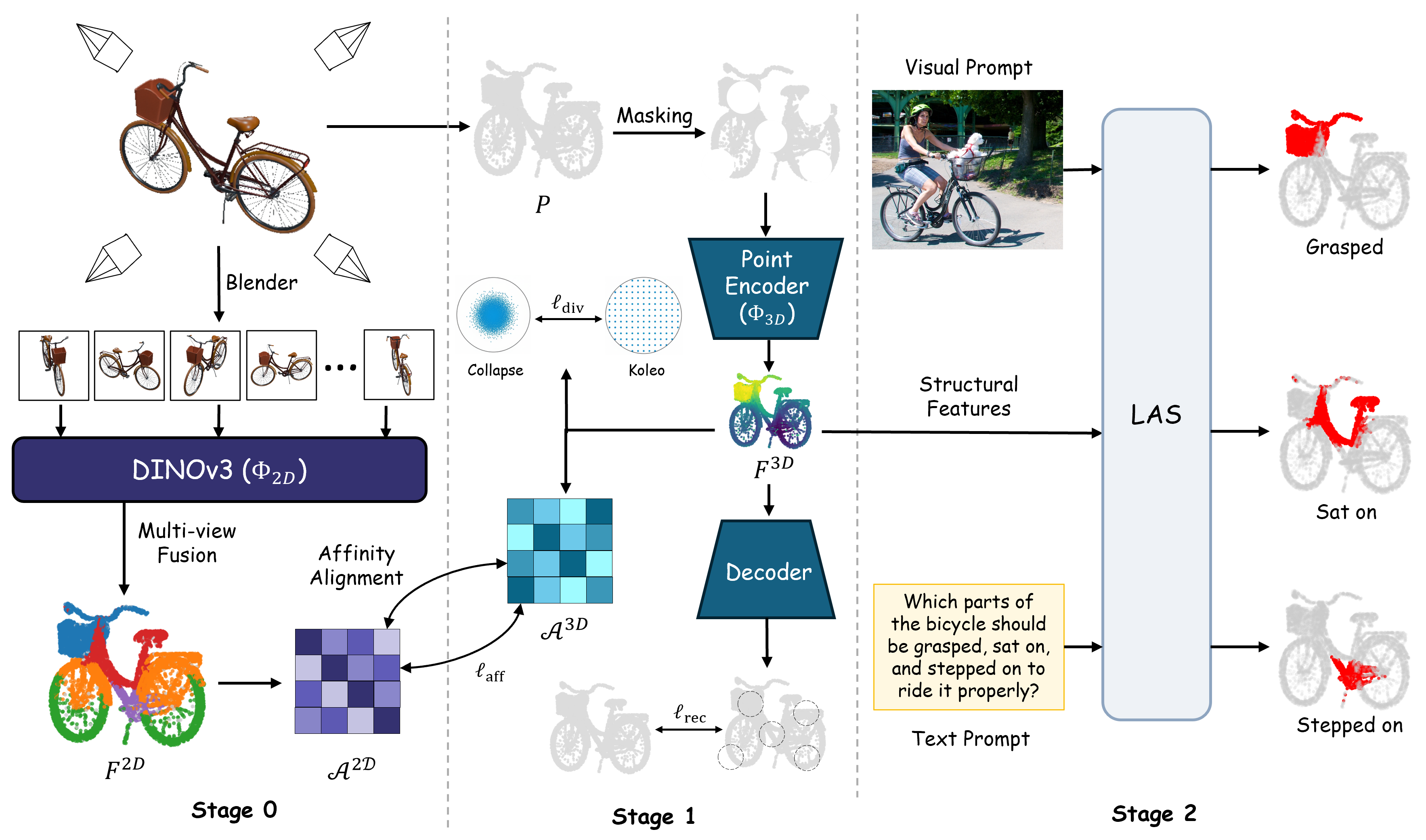}
\caption{Overview of our three-stage framework for prompt-guided 3D affordance segmentation. Stage 0 (2D Semantic Knowledge Extraction) associates each point cloud $P$ with multi-view 2D features extracted by a frozen encoder $\Phi_{2D}$, producing lifted per-point semantic knowledge $F^{2D}$ and affinity matrix $A^{2D}$. Stage 1 (Cross-Modal Affinity Transfer) pretrains the 3D backbone $\Phi_{3D}$ by aligning the affinity matrix of 3D features $A^{3D}$ with the corresponding 2D affinity matrix $A^{2D}$. 
Stage 2 utilizes the Lightweight Affordance Segmentor (LAS) to fine-tune $\Phi_{3D}$ with multi-modal prompts (textual or visual) to generate the final prompt-conditioned affordance map $\mathbf{M}$.}

    \label{fig:p3}
    \vspace{-2mm}
\end{figure*}

\section{Methodology}

Our approach builds a unified framework for 3D affordance segmentation by progressively introducing semantic knowledge into 3D representations and adapting them to prompt-driven affordance segmentation task. The entire process involves three tightly connected stages that move from cross-modal grounding to task-specific adaptation.

\subsection{Overview.}

Given an input 3D point cloud $P=\{\mathbf{p}_i\in\mathbb{R}^3\}_{i=1}^N$ with $N$ points, our framework's goal is to produce a dense, prompt-conditioned affordance map $\mathbf{M}\in\mathbb{R}^N$. The overall process consists of three progressive stages. 
In Stage 0 (2D Semantic Knowledge Extraction), we pre-process 3D objects to obtain per-point 2D semantic descriptors, $F^{2D}=\{\mathbf{f}^{2D}_i\in\mathbb{R}^{d_{2D}}\}_{i=1}^N$, which serve as the foundational supervision signal. 
In Stage 1 (Cross-Modal Affinity Transfer, CMAT), we leverage these $F^{2D}$ features to pretrain our 3D backbone $\Phi_{3D}$. This stage results in a 3D model that internalizes a functionally structured 3D representation by aligning 3D patch affinity with 2D patch affinity.
Finally, in Stage 2 (Lightweight Affordance Segmentor), this pretrained backbone $\Phi_{3D}$ is integrated with a multi-modal fusion module, which accepts $P$ and a text or visual prompt to output the final segmentation map $\mathbf{M}$.

\subsection{\textbf{Stage 0: 2D Semantic Knowledge Extraction}}

To establish transferable semantic knowledge for 3D learning, we prepare a dataset of 3D objects paired with multi-view 2D features. The dataset includes over 10,000 3D models from Objaverse~\cite{objaverse} and Behavior-1K~\cite{behavior1k}, covering 101 everyday object categories such as furniture, kitchenware, and tools, ensuring rich semantic coverage.

Each 3D model is represented as a point cloud $P={\mathbf{p}i\in\mathbb{R}^3}_{i=1}^N$. We render $V$ RGB views under uniformly distributed camera poses and process them using a frozen 2D encoder $\Phi_{2D}$ (DINOv3~\cite{dinov3}) to obtain dense feature maps. This multi-view setup ensures that both visible and previously occluded surfaces receive consistent 2D semantic cues. These per-pixel embeddings are back-projected and interpolated onto the corresponding 3D points following established feature lifting techniques~\cite{d3field}, producing the per-point semantic descriptors:
\[
F^{2D}=\{\mathbf{f}^{2D}_i\in\mathbb{R}^{d_{2D}}\}_{i=1}^N.
\]
These lifted features act as a high-quality semantic grounding signal that guides the subsequent CMAT pretraining stage.

\subsection{\textbf{Stage 1: Cross-Modal Affinity Transfer}}

Stage 0 provides semantic knowledge for each point, but we still need a training strategy to enable the 3D backbone $\Phi_{3D}$ to internalize this structure. Unlike self-supervised methods~\cite{qi2017pointnetplusplus,pointmae} that focus mainly on geometric reconstruction, our goal is not only to preserve geometric continuity but also to infuse the backbone with functional structure derived from the 2D domain. We accomplish this through our \textbf{Cross-Modal Affinity Transfer (CMAT)} scheme.

The backbone follows a PointMAE-style~\cite{pointmae} transformer encoder, processing the point cloud as patch tokens $P_S = \{P_j\}_{j=1}^m$. For each patch, we obtain patch-level 2D and 3D features by average pooling:
\begin{align}
    \bar{\mathbf{f}}^{2D}_j &= \frac{1}{|P_j|}\sum_{\mathbf{p}_i \in P_j}\mathbf{f}^{2D}_i, \\
    \bar{\mathbf{f}}^{3D}_j &= \frac{1}{|P_j|}\sum_{\mathbf{p}_i \in P_j}\mathbf{f}^{3D}_i,
\end{align}
where $\mathbf{f}^{2D}_i$ is the lifted semantic feature from Stage 1 and $\mathbf{f}^{3D}_i$ is the output of $\Phi_{3D}$.

We then construct cross-modal affinity matrices to capture relational structure. The teacher affinity matrix $\mathcal{A}^{2D}$ is defined as:
\begin{equation}
    \mathcal{A}^{2D}_{jk} = \frac{\bar{\mathbf{f}}^{2D}_j \cdot \bar{\mathbf{f}}^{2D}_k}{\|\bar{\mathbf{f}}^{2D}_j\|\ \|\bar{\mathbf{f}}^{2D}_k\|},
\end{equation}
and $\mathcal{A}^{3D}$ is computed analogously from $\bar{\mathbf{f}}^{3D}_j$. Here, $m$ is the total number of patches, and the indices $j, k \in \{1, \dots, m\}$ iterate over all patch pairs. These matrices encode part–whole semantic relationships in two modalities, with the 2D space providing structural guidance for the 3D encoder.

The pretraining objective is centered on our proposed semantic alignment loss, supported by two auxiliary losses for geometric stability and feature diversity.

\begin{figure*}[ht]
    \centering
    \includegraphics[width=1.0\linewidth]{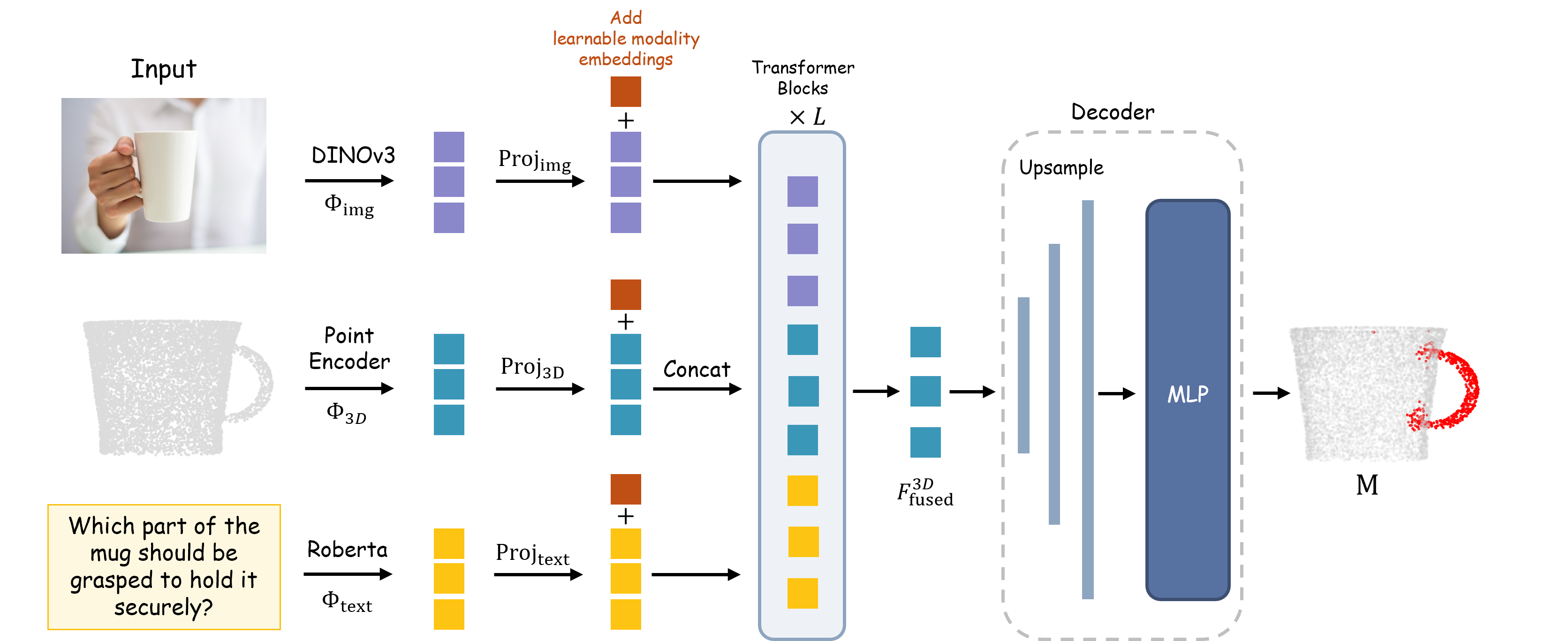}
\caption{Architecture of the Lightweight Affordance Segmentor. This module fuses geometric patch tokens from our pretrained 3D backbone with multi-modal prompts (text and/or visual) in a shared embedding space. A stack of co-attentional Transformer blocks enables bidirectional interaction between geometric features and prompt tokens, leading to prompt-conditioned 3D understanding. The resulting patch features are then upsampled to per-point resolution to generate the final affordance mask.}

    \label{fig:p4} 
\end{figure*}

\paragraph{Semantic Alignment ($\ell_{\mathrm{aff}}$).}
Our primary objective is to transfer the structured functional relationships from the 2D teacher space. We introduce the Semantic Alignment Loss ($\ell_{\mathrm{aff}}$) to align the affinity matrix produced by the 3D student encoder ($\mathcal{A}^{3D}$) with the teacher affinity matrix ($\mathcal{A}^{2D}$). This forces the 3D feature space to reflect the same inter-part semantic relations encoded in the 2D space, injecting functional structure without requiring explicit semantic labels or part annotations:
\begin{equation}
    \ell_{\mathrm{aff}} = \frac{1}{m^2}\sum_{j=1}^{m}\sum_{k=1}^{m}\bigl(\mathcal{A}^{3D}_{jk}-\mathcal{A}^{2D}_{jk}\bigr)^2.
\end{equation}

To maintain capability of geometry reconstruction, we adopt two established auxiliary objectives. First, to maintain the underlying structure of the 3D shape, we employ a geometric fidelity loss ($\ell_{\mathrm{rec}}$) based on the masked autoencoding strategy from PointMAE~\cite{pointmae}, which preserves the backbone’s ability for geometric reconstruction and understanding of point clouds. Second, to prevent feature collapse and ensure the embeddings are well-separated and expressive, we apply a feature diversity loss ($\ell_{\mathrm{div}}$). We use the KoLeo regularizer~\cite{koloe} for this purpose, which penalizes small nearest-neighbor distances in the embedding space to maximize feature entropy.

The final pretraining loss is a weighted sum of these components:
\begin{equation}
    \ell_{\mathrm{pretrain}} = \lambda_{\mathrm{aff}}\ell_{\mathrm{aff}} + \lambda_{\mathrm{rec}}\ell_{\mathrm{rec}} + \lambda_{\mathrm{div}}\ell_{\mathrm{div}}.
\end{equation}

\subsection{\textbf{Stage 2: Lightweight Affordance Segmention}}

In the final stage, we adapt the CMAT-pretrained backbone for the prompt-guided affordance segmentation task. We employ a lightweight segmentation transformer based on a standard co-attentional architecture to efficiently fuse geometric and prompt features.

The segmentor first processes the input point cloud using the pretrained encoder $\Phi_{3D}$ to obtain geometric patch tokens $F^{3D}$. User-provided prompts are encoded into feature vectors: a textual phrase (e.g.\textit{``Which part of the mug should be grasped''}) by $\Phi_{\mathrm{text}}$ to $F_{\mathrm{text}}$, and a visual exemplar by $\Phi_{\mathrm{img}}$ to $F_{\mathrm{img}}$. 
To enable cross-modal interaction, these features are projected into a shared embedding space, with learnable modality embeddings added to preserve their source identity:
\begin{align}
    \mathbf{T}_P &= \mathrm{Proj}_{3D}(F^{3D}) + \mathbf{E}_{\mathrm{point}}, \\
    \mathbf{T}_{\mathrm{text}} &= \mathrm{Proj}_{\mathrm{text}}(F_{\mathrm{text}}) + \mathbf{E}_{\mathrm{text}}, \\
    \mathbf{T}_{\mathrm{img}}  &= \mathrm{Proj}_{\mathrm{img}}(F_{\mathrm{img}}) + \mathbf{E}_{\mathrm{img}}.
\end{align}

\begin{figure*}[ht]
    \centering
    \includegraphics[width=1.0\linewidth]{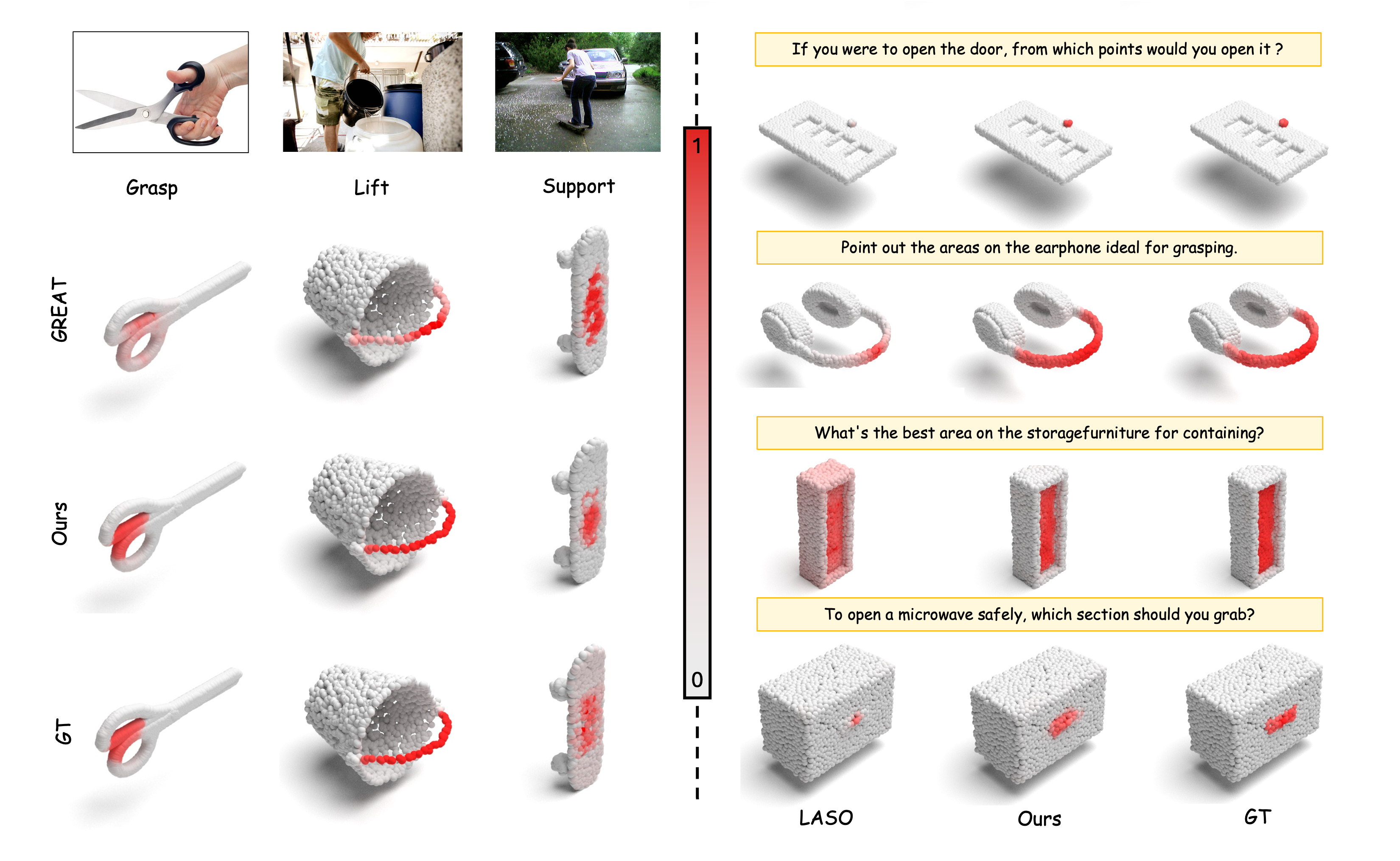}
    \caption{Qualitative comparison on challenging cases from the PIADv2~\cite{great} (visual prompt) and LASO~\cite{LASO} (text prompt) datasets. These examples visually corroborate our quantitative improvements and highlight our framework's superior fine-grained segmentation capability.}
    \label{fig:qua}
    \vspace{-2mm}
\end{figure*}
All available prompt tokens (e.g., $\mathbf{T}_{\mathrm{text}}$, $\mathbf{T}_{\mathrm{img}}$, or both) are aggregated into a set $\mathbf{T}_Q$ and concatenated with the geometric tokens to form the fused sequence $[\mathbf{T}_Q;\mathbf{T}_P]$. This sequence is processed by a stack of $L$ co-attentional transformer blocks. The self-attention mechanism facilitates deep, bidirectional interaction, allowing geometric tokens to be conditioned by prompts and prompts to ground in the 3D geometry.

Finally, the resulting prompt-conditioned patch features, $F^{3D}_{\mathrm{fused}}$, are upsampled to the original point resolution via feature propagation. A lightweight MLP head then maps these per-point features to the final segmentation logits $\mathbf{M}$.

%% file: sec/4_experiment.tex
\section{Experiments}
\label{sec:experiments}
In this section, we conduct a series of experiments to thoroughly evaluate our proposed three-stage framework. We first introduce the experimental setup, including the datasets and evaluation metrics. We then detail our implementation for reproducibility. Subsequently, we present quantitative comparisons against state-of-the-art methods on both visual and text-prompted affordance segmentation tasks. Finally, we perform extensive ablation studies to analyze the contribution of each key component in our framework and showcase qualitative results to provide intuitive insights.

It is important to note that our framework is inherently designed to process multi-modal prompts, such as a combination of visual cues and textual instructions. However, existing benchmarks are constrained to single-modality inputs, supporting either visual or text prompts but not their concurrent use. To ensure a fair comparison on these datasets, we adapt our model by providing a \texttt{null} input for the modality not supported by the specific benchmark. We posit that the full capabilities of our method will be even more evident on future benchmarks designed for true multi-modal queries.

\subsection{Experimental Setup}
\paragraph{Datasets.}
We conduct our evaluation on several established benchmarks for affordance segmentation. We utilize the Point Image Affordance Dataset (PIAD)~\cite{Yang_2023_ICCV} and its large-scale extension, PIADv2~\cite{great}. PIAD contains 5162 interaction images and 7012 3D objects across 23 categories, while PIADv2 significantly increases this scale to approximately 15213 images and 38889 3D instances spanning 43 object categories. These datasets provide diverse object interactions and affordance types, covering both simple tools and complex articulated categories. Additionally, we employ the Language-guided Affordance Segmentation on 3D Objects (LASO) dataset~\cite{LASO}, which consists of 19751 point-question pairs covering 8434 object shapes. LASO further evaluates a model’s ability to understand natural-language queries and generalize across heterogeneous object geometries. For all datasets, we strictly adhere to their official train/test splits and Seen/Unseen splits to ensure fair comparison.

\paragraph{Evaluation Metrics.}
We assess segmentation quality using six metrics. We evaluate performance on average IoU (aIoU)~\cite{iou}, confidence ranking (AUC)~\cite{Lobo2008AUCAM}, shape fidelity (SIM)~\cite{sim}, and score error (MAE)~\cite{mae}. This comprehensive suite enables a multi-faceted evaluation beyond simple segmentation overlap.

\begin{table*}[!t]
    \begin{center}
        \small
        \caption{\textbf{Quantitative comparison with state-of-the-art methods on the PIAD and PIADv2 datasets.} We report results on both Seen and Unseen splits. The \textbf{best} and \underline{second-best} results are highlighted. \textbf{Ours(w/o CMAT)} denotes our LAS segmentor paired with the standard PointMAE backbone~\cite{pointmae}, using its original pretrained weights.}
        \label{tab:styled_combined_results}
        \setlength{\tabcolsep}{1.35pt} 
        \begin{tabular}{lcccccccccccccccc}
            \toprule
            
             & \multicolumn{8}{c}{\textbf{PIAD}} & \multicolumn{8}{c}{\textbf{PIADv2}} \\
             \cmidrule(lr){2-9} \cmidrule(lr){10-17}
             & \multicolumn{4}{c}{Seen} & \multicolumn{4}{c}{Unseen} & \multicolumn{4}{c}{Seen} & \multicolumn{4}{c}{Unseen} \\
             \cmidrule(lr){2-5} \cmidrule(lr){6-9} \cmidrule(lr){10-13} \cmidrule(lr){14-17}
             Method & AUC $\uparrow$ & aIoU $\uparrow$ & SIM $\uparrow$ & MAE $\downarrow$ & AUC $\uparrow$ & aIoU $\uparrow$ & SIM $\uparrow$ & MAE $\downarrow$ & AUC $\uparrow$ & aIoU $\uparrow$ & SIM $\uparrow$ & MAE $\downarrow$ & AUC $\uparrow$ & aIoU $\uparrow$ & SIM $\uparrow$ & MAE $\downarrow$ \\
            \midrule
            
            \multicolumn{17}{c}{\cellcolor{cell}\textit{\textbf{State-of-the-art Methods}}} \\
            MBDF~\cite{mbdf}        & 74.9 & 9.3 & 0.415 & 0.143 & 58.2 & 4.2 & 0.325 & 0.213 & ---   & ---   & ---   & ---   & ---   & ---   & ---   & ---   \\
            PMF~\cite{epmf}         & 75.1 & 10.1 & 0.425 & 0.141 & 60.3 & 4.7 & 0.330 & 0.211 & ---   & ---   & ---   & ---   & ---   & ---   & ---   & ---   \\
            FRCNN~\cite{frcnn}       & 76.1 & 12.0 & 0.429 & 0.136 & 61.9 & 5.1 & 0.332 & 0.195 & 87.05 & 33.55 & 0.600 & 0.082 & 72.20 & 18.08 & 0.362 & 0.152 \\
            ILN~\cite{iln}        & 75.8 & 11.5 & 0.427 & 0.137 & 59.7 & 4.7 & 0.325 & 0.207 & ---   & ---   & ---   & ---   & ---   & ---   & ---   & ---   \\
            PFusion~\cite{pfusion}     & 77.5 & 12.3 & 0.432 & 0.135 & 61.9 & 5.3 & 0.330 & 0.193 & ---   & ---   & ---   & ---   & ---   & ---   & ---   & ---   \\
            XMF~\cite{xmf}         & 78.2 & 12.9 & 0.441 & 0.127 & 62.6 & 5.7 & 0.342 & 0.188 & 87.39 & 33.91 & 0.604 & 0.078 & 74.61 & 17.40 & 0.361 & 0.126 \\
            IAGNet~\cite{Yang_2023_ICCV}     & 84.9 & \underline{20.5} & 0.545 & \underline{0.098} & \underline{71.8} & \underline{8.0} & 0.352 & 0.127 & 89.03 & 34.29 & 0.623 & 0.076 & 73.03 & 16.78 & 0.351 & \underline{0.123} \\
            LASO~\cite{LASO}   & 84.2 & 19.7 & 0.590 & \textbf{0.096} & 69.2 & \underline{8.0} & \underline{0.386} & \textbf{0.118} & 90.34 & 34.88 & 0.627 & 0.077 & 73.32 & 16.05 & 0.354 & \underline{0.123} \\
            GREAT~\cite{great}       & ---  & ---  & ---   & ---   & ---  & ---  & ---   & ---   & \underline{91.99} & 38.03 & 0.676 & \textbf{0.067} & \underline{79.57} & 20.16 & 0.402 & \textbf{0.109} \\
            \midrule
            
            \multicolumn{17}{c}{\cellcolor{cell}\textit{\textbf{Our Models}}} \\
            Ours(w/o CMAT)    & \underline{85.5} & 19.9 & \underline{0.621} & 0.105 & 71.0 & 7.8 & 0.379 & 0.130 & 91.45 & \underline{38.16} & \underline{0.752} & 0.089 & 76.87 & \underline{22.07} & \underline{0.542} & 0.142 \\
            \textbf{Ours(w/CMAT)} & \textbf{88.0} & \textbf{21.4} & \textbf{0.725} & 0.099 & \textbf{74.8} & \textbf{8.4} & \textbf{0.426} & \underline{0.125} & \textbf{93.01} & \textbf{44.88} & \textbf{0.783} & \underline{0.071} & \textbf{83.95} & \textbf{27.40} & \textbf{0.593} & 0.127 \\
            \bottomrule
        \end{tabular}
    \end{center}
    \vspace{-5mm}
\end{table*}

\begin{table}[t]
    \centering
    \small
    \caption{The overall results of all comparative methods on the \textbf{LASO} dataset. \textbf{Seen} and \textbf{Unseen} are two partitions of the dataset. The \textbf{best} and \underline{2nd best} scores from each metric are highlighted in \textbf{bold} and \underline{underlined}, respectively.}
    \vspace{-0.2cm}
    \resizebox{\linewidth}{!}{
    \begin{tabular}{c|c|cccc}
    \toprule
    \textbf{Type} & \textbf{Method} & \textbf{aIoU} $\uparrow$ & \textbf{AUC} $\uparrow$ & \textbf{SIM} $\uparrow$ & \textbf{MAE} $\downarrow$ 
    \\
    \midrule
    \multirow{8}{*}{\textbf{Seen}} 
    & ReferTrans~\cite{refertrans} & $13.7$ & $79.8$ & $0.497$ & $0.124$ 
    \\
    & ReLA~\cite{rela} & $15.2$ & $78.9$ & $0.532$ & $0.118$ 
    \\
    & 3D-SPS~\cite{3dsps}  & $11.4$ & $76.2$ & $0.433$ & $0.138$ 
    \\
    & IAGNet~\cite{Yang_2023_ICCV} & $17.8$ & $82.3$ & $0.561$ & $0.109$
    \\
    & LASO~\cite{LASO} & \underline{$20.8$} & \underline{$87.3$} & \underline{$0.629$} & \textbf{0.093} 
    \\
    & \cellcolor{cell}\textbf{Ours} & \cellcolor{cell}\textbf{21.7} & \cellcolor{cell}\textbf{88.2} & \cellcolor{cell}\textbf{0.681} & \cellcolor{cell}\underline{0.096}
    \\ 
    \midrule
    \multirow{8}{*}{\textbf{Unseen}} 
    & ReferTrans~\cite{refertrans} & $10.2$ & $69.1$ & $0.432$ & $0.145$ 
    \\
    & ReLA~\cite{rela} & $10.7$ & $69.7$ & $0.429$ & $0.144$ 
    \\
    & 3D-SPS~\cite{3dsps} & $7.9$ & $68.8$ & $0.402$ & $0.158$ 
    \\
    & IAGNet~\cite{Yang_2023_ICCV} & $12.9$ & $77.8$ & $0.443$ & $0.129$ 
    \\
    & LASO~\cite{LASO} & $14.6$ & \underline{$80.2$} & $\underline{0.507}$ & $0.119$ 
    \\
    & \cellcolor{cell}\textbf{Ours} & \cellcolor{cell}\textbf{17.5} & \cellcolor{cell}\textbf{82.9} & \cellcolor{cell}\textbf{0.625} & \cellcolor{cell}\underline{0.112}
    \\ 
    \bottomrule
    \end{tabular}}
    \vspace{-5mm}
\label{tab:laso_table}
\end{table}

\subsection{Implementation Details}

\paragraph{Stage 0: 2D Semantic Knowledge Extraction.}
The 2D teacher model, $\Phi_{2D}$, is a pretrained DINOv3 with a ViT-Large backbone, with its weights kept frozen throughout the process. For each 3D object in our pretraining dataset, we render $V=12$ RGB views at a resolution of 224x224. These views are rendered from camera poses uniformly distributed around the object to maximize the coverage of its visible surface. We extract dense features from the final layer of $\Phi_{2D}$ and lift them back to the 3D point cloud using an inverse projection and nearest-neighbor interpolation, resulting in the per-point semantic feature set $F^{2D}$.

\paragraph{Stage 1: Cross-Modal Affinity Transfer (CMAT).}
Our 3D backbone, $\Phi_{3D}$, is a PointMAE-style transformer encoder with $12$ blocks and an embedding dimension of $384$. The input point cloud is grouped into $64$ patches. For the geometric reconstruction task, we employ a high masking ratio of $60\%$. The weights for the combined pretraining objective are set to $\lambda_{\mathrm{rec}}=1.0$, $\lambda_{\mathrm{aff}}=0.1$, and $\lambda_{\mathrm{div}}=0.2$. We pretrain the model for $150$ epochs with a batch size of $128$ using the AdamW optimizer. The learning rate starts at $1\text{e-}4$ and decays following a cosine schedule with a warmup period of $15$ epochs.

\paragraph{Stage 2: Lightweight Affordance Segmentor.}
For prompt encoding, the text encoder $\Phi_{\mathrm{text}}$ is a pretrained RoBERTa-base model~\cite{roberta}. The visual prompt encoder $\Phi_{\mathrm{img}}$ is a pretrained DINOv3 with a ViT-B backbone. The weights of both prompt encoders are kept frozen during fine-tuning. The affordance segmentor is built with $L=6$ co-attentional fusion blocks. All input point clouds for downstream tasks are uniformly sampled to $2048$ points. During fine-tuning, we employ a differential learning rate: the pretrained backbone $\Phi_{3D}$ uses a lower learning rate of $1\text{e-}5$, while the newly initialized modules in segmentor and segmentation head use a higher rate of $1\text{e-}4$. The weights for the segmentation loss are $\lambda_{\mathrm{focal}}=1.0$ and $\lambda_{\mathrm{dice}}=1.0$. Each model is fine-tuned for $100$ epochs with a batch size of $16$. All three stages are conducted on $4$ NVIDIA RTX 3090 GPUs.


\subsection{Quantitative Results}

\paragraph{Evaluation on PIAD~\cite{Yang_2023_ICCV} and PIADv2~\cite{great}.}
Table~\ref{tab:styled_combined_results} summarizes the quantitative comparison on the visual-prompted affordance segmentation benchmarks. 
On PIAD, our method achieves a SOTA performance, particularly excelling in maintaining fine structural consistency. 
The SIM score reaches 0.725, marking a 22.9\% relative improvement over the previous best. 
The advantage becomes even clearer on PIADv2, a dataset designed with higher visual complexity and denser affordance categories. 
Our model outperforms the prior top performer by 7.85 aIoU points on the Seen split and 5.33 points on the Unseen Object split, 
demonstrating not only strong recognition of seen objects but also reliable generalization to novel categories. 
These results indicate that aligning 3D representations with semantic structure through CMAT fundamentally strengthens affordance reasoning beyond geometric similarity.

\paragraph{Evaluation on LASO.}
We further evaluate on LASO, where segmentation is guided purely by textual instructions. 
As shown in Table~\ref{tab:laso_table}, our framework achieves the highest aIoU on both the Seen and Unseen splits, reaching 21.7\% and 17.5\%, respectively. 
Unlike previous works that struggle to connect linguistic cues to precise 3D regions, our approach consistently grounds language into meaningful geometric contexts. 
This suggests that the learned feature space after CMAT pretraining provides a stable semantic foundation, allowing even short prompts to activate correct functional regions in complex 3D structures.

\paragraph{Model Efficiency Analysis.}
A key strength of our framework lies in its efficiency–performance trade-off. 
As summarized in Table~\ref{tab:model_param_vram_aiou}, our model maintains a modest footprint of 300M parameters and 4--8 GB VRAM, 
comparable to conventional non-MLLM architectures. 
In contrast, recent MLLM-based methods such as GREAT~\cite{great} require more than 4B parameters and up to 30 GB of memory. 
Despite being nearly an order of magnitude smaller, our model achieves 44.88 aIoU on PIADv2, outperforming all previous methods. 
This balance between scalability and precision highlights the practicality of our framework for real-world affordance understanding systems.

\begin{table}[H]
\centering
\small
\setlength{\tabcolsep}{7pt}
\caption{Comparison of model efficiency. aIoU is evaluated on the PIADv2 seen split. $^{\dagger}$ denotes the use of an MLLM backbone.}
\begin{tabularx}{0.45\textwidth}{lccc}
\toprule
Model & Params & VRAM & aIoU \\
\midrule
IAGNet (ICCV23) & 30M & 0.8--2.0 GB & 34.29 \\
LASO (CVPR24) & 130M & 2.0--3.5 GB & 34.88 \\
GREAT$^{\dagger}$ (CVPR25) & 4B & 16--30 GB & 38.03 \\
Ours & 300M & 4--8 GB & \textbf{44.88} \\
\bottomrule
\end{tabularx}

\label{tab:model_param_vram_aiou}
\end{table}

\subsection{Ablation Studies}

To better understand each component’s contribution, we perform detailed ablation studies on the PIADv2 Seen split, summarized in Table~\ref{tab:ablation_final}. 
We begin with the pretraining objectives in Stage~1. 
A model trained with only the reconstruction loss ($\mathcal{L}_{\mathrm{rec}}$) reaches 39.27\% aIoU, indicating that geometric signals alone are insufficient for functional reasoning. Increasing the amount of training data can slightly enhance the backbone’s representational ability, but this improvement does not lead to a fundamental change.
Introducing the core affinity alignment loss ($\mathcal{L}_{\mathrm{aff}}$) yields a significant jump to 44.13\%, confirming that aligning cross-modal feature affinities is central to the success of CMAT. 
Adding the feature diversity regularization ($\mathcal{L}_{\mathrm{div}}$) further improves the score to 44.88\%, indicating that encouraging local feature variation helps capture fine-grained affordance boundaries. 
We also observe that a stronger 2D teacher (DINOv3 vs. DINOv2) consistently enhances downstream results, supporting the idea that richer 2D semantics transfer more effectively into 3D. 

For the segmentation backbone, we compare PointNet++ and PointMAE trained from scratch. 
Although PointMAE performs slightly better, both lag far behind our CMAT-pretrained model, which reaches 44.88\%. 
This clear margin demonstrates that CMAT provides not just better initialization, but a fundamentally stronger semantic organization of 3D space.

\begin{table}[ht]
    \centering
    \caption{Ablation studies on the PIADv2 Seen split. We analyze Stage 2 pretraining components and Stage 3 backbone choices.}
    \label{tab:ablation_final}
    \resizebox{\columnwidth}{!}{%
    \begin{tabular}{llc}
        \toprule
        \multicolumn{2}{l}{Configuration} & aIoU (\%) $\uparrow$ \\
        \midrule
        \multicolumn{3}{l}{\cellcolor{cell}\textit{Stage 1: pretraining Objective (on PointMAE with DINOv3 Teacher)}} \\
        (1) & $\mathcal{L}_{\mathrm{rec}}$ only & 39.27 \\
        (2) & $\mathcal{L}_{\mathrm{rec}} + \mathcal{L}_{\mathrm{aff}}$ & 44.13 \\
        (3) & Full Objective ($\mathcal{L}_{\mathrm{rec}} + \mathcal{L}_{\mathrm{aff}} + \mathcal{L}_{\mathrm{div}}$) & \textbf{44.88} \\
        (4) & Full Objective (Replacing DINOv3 with DINOv2) & 43.26 \\
        \midrule
        \multicolumn{3}{l}{\cellcolor{cell}\textit{Stage 2: Segmentor Backbone Choices}} \\
        (5) & PointNet++ Backbone & 37.91 \\
        (6) & PointMAE Backbone & 38.16 \\
        \bottomrule
    \end{tabular}%
    }
    \vspace{-5mm}
\end{table}

\subsection{Qualitative Analysis}

Figure~\ref{fig:qua} visualizes qualitative comparisons on PIADv2 and LASO, focusing on challenging, fine-grained interactions. 
Given a visual prompt showing a hand grasping scissors, our model precisely isolates the handle region, while the baseline misidentifies nearby metallic edges. 
Similarly, when given the text prompt ``open the door'', our model highlights the entire handle area with accurate boundaries, whereas others capture only partial fragments. 
Across diverse categories, our predictions show consistent localization of functionally relevant parts, reflecting a nuanced understanding of how objects afford actions. 
These visual findings indicate that CMAT enables representations that respect both geometry and purpose, leading to more human-like affordance perception in 3D.

%% file: sec/5_conclusion.tex
\section{Conclusion}

In this work, we address the inherent semantic ambiguity of 3D point clouds by proposing a novel paradigm that grounds 3D representations in the rich knowledge of 2D Vision Foundation Models. Our method is centered on a Cross-Modal Affinity Transfer (CMAT) pretraining strategy, which teaches a 3D encoder to learn a structurally superior feature space, later adapted for affordance segmentation tasks by our lightweight affordance segmentor. Our approach achieves state-of-the-art performance, with comprehensive ablations confirming that these significant gains are directly attributable to our CMAT strategy's effective transfer of relational knowledge across modalities. Ultimately, this work presents a generalizable and powerful paradigm for injecting 2D semantic knowledge into 3D feature learning, holding significant promise for a wide range of future 3D understanding tasks.

\section{Acknowledgment}
This work was supported by the NSFC under Grant 62322604 and 62576207.